\documentclass[journal]{IEEEtranTIE}
\usepackage{graphicx}
\usepackage{cite}
\usepackage{picinpar}
\usepackage{amsmath}
\usepackage{url}
\usepackage{flushend}
\usepackage[latin1]{inputenc}
\usepackage{colortbl}
\usepackage{soul}
\usepackage{multirow}
\usepackage{pifont}
\usepackage{color}
\usepackage{alltt}
\usepackage[hidelinks]{hyperref}
\usepackage{enumerate}
\usepackage{siunitx}
\usepackage{breakurl}
\usepackage{epstopdf}
\usepackage{pbox}
\usepackage{amsfonts,amssymb} 
\usepackage{bbm}
\usepackage{amsmath}
\usepackage{graphicx}
\usepackage{subfigure}
\usepackage{graphicx,times}
\usepackage{subfigure}         
\usepackage{array}
\usepackage{etoolbox}
\usepackage{cite}
\usepackage{bm}
\usepackage{multirow}
\usepackage{array}
\usepackage{hyperref}
\hypersetup{hidelinks,
	colorlinks=true,
	allcolors=black,
	pdfstartview=Fit,
	breaklinks=true}
 
\begin{document}
\title{	Nowhere to Go: Benchmarking Multi-robot Collaboration in Target Trapping Environment }

\author{
	\vskip 1em
	
	Hao Zhang, Jiaming Chen, Jiyu Cheng, Yibin Li, Simon X. Yang, \IEEEmembership{Senior Member, IEEE}, and Wei Zhang, \IEEEmembership{Senior Member, IEEE}

	\thanks{
            Manuscript received Month xx, 2xxx; revised Month xx, xxxx; accepted
Month x, xxxx. This work is supported by xxxx.(Hao Zhang and Jiaming Chen contributed equally to this work.) (Corresponding author: Jiyu Cheng.)

		Hao Zhang, Jiaming Chen, Jiyu Cheng, Yibin Li, and Wei Zhang are with the School of Control Science and Engineering, Shandong University, Jinan 250061, China (e-mail: hao.zhang@mail.sdu.edu.cn; jamin@mail.sdu.edu.cn; {\{jycheng, liyb, davidzhang\}@sdu.edu.cn}).
		
		Simon X. Yang is with the Advanced Robotics and Intelligent Systems
Laboratory, School of Engineering, University of Guelph, Guelph, ON
N1G2W1, Canada (e-mail: syang@uoguelph.ca).
	}
}

\maketitle
	
\begin{abstract}

Collaboration is one of the most important factors in multi-robot systems. Considering certain real-world applications and to further promote its development, we propose a new benchmark to evaluate multi-robot collaboration in Target Trapping Environment (T2E). In T2E, two kinds of robots (called \emph{captor robot} and \emph{target robot}) share the same space. The captors aim to catch the target collaboratively, while the target will try to escape from the trap. Both the trapping and escaping process can use the environment layout to help achieve the corresponding objective, which requires high collaboration between robots and the utilization of the environment.
For the benchmark, we present and evaluate multiple learning-based baselines in T2E, and provide insights into regimes of multi-robot collaboration. We also make our benchmark publicly available and encourage researchers from related robotics disciplines to propose, evaluate, and compare their solutions in this benchmark. Our project is released at \href{https://github.com/Dr-Xiaogaren/T2E}{https://github.com/Dr-Xiaogaren/T2E}.

\end{abstract}


%
{}

\definecolor{limegreen}{rgb}{0.2, 0.8, 0.2}
\definecolor{forestgreen}{rgb}{0.13, 0.55, 0.13}
\definecolor{greenhtml}{rgb}{0.0, 0.5, 0.0}

\section{Introduction}

\IEEEPARstart{M}{ulti}-robot hunting, which is also called predator-prey or pursuit-evasion, is a classical but challenging task, which requires multiple hunter robots to chase a prey robot in an enclosed space. In this process, if the prey robot falls in the capture range of a certain predator robot, then the prey robot is considered to be successfully caught. Due to its wide applications, the task has been well studied in past decades and some remarkable methods have been proposed\cite{c38,c39,c40}. However, in some real-world scenarios such as military tasks like fighter jet interception\cite{c36} and fleet confrontation\cite{c37}, it is more practical for the hunter robots to encircle the prey robot to trap it other than catching it by just one hunter at some time point. And the trap can be achieved by the robot team or robots and the environmental elements like obstacles. 

\begin{figure}
  \centering

  \includegraphics[width=8.65cm,height=4.867cm]{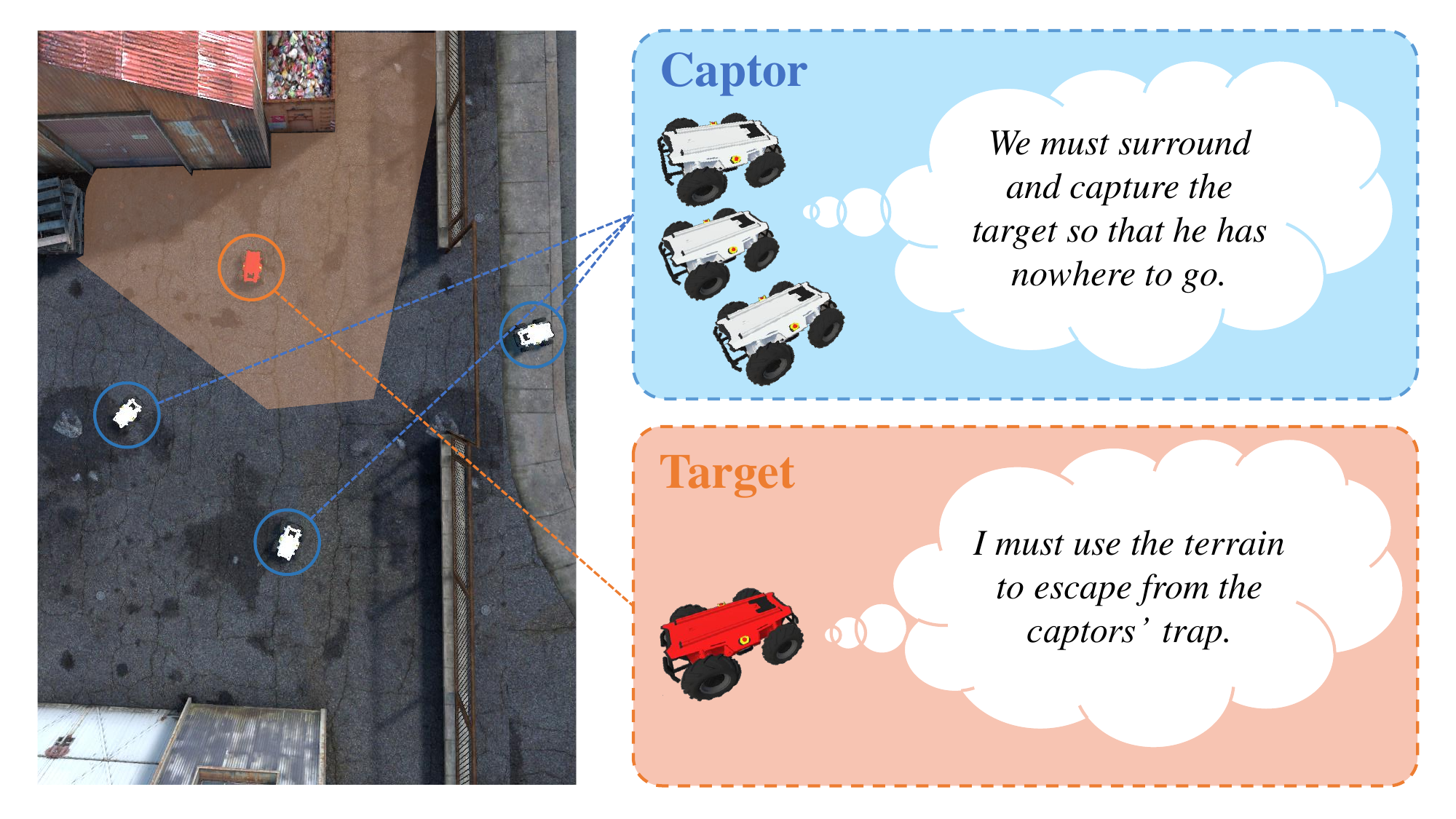}
  \caption{Illustration of the multi-robot target trapping task. In this task, captor robots need to surround the target robot so that it cannot move. The red mask area is the Absolutely Safe Zone of the target robot, which represents its degree of oppression.}
  \label{fig:figure-1}
  \vspace{-0.3cm}
\end{figure}

To facilitate the research on the above mentioned scenarios, in this paper we propose the multi-robot target trapping task and establish the target trapping environment to better align with real-world requirements and attempt to well stimulate collaboration among the robot team or robots with the environment. Different from the hunting task, target trapping requires multiple captor robots to utilize the collaboration and the environment layout to surround the target robot until it cannot move. The captor robots can choose to encircle the target robot or force it into a corner. Meanwhile, the target robot can also use obstacles in the environment to escape from the captor robots. More concretely, it requires captor robots to generate a trap with the environment to limit the target robot in a small area and ultimately leave it with no way to escape, which demands effective collaboration between robots and the environment. Furthermore, we propose the concept of Absolutely Safe Zone to describe the constraint level of the target robot and provide a rigorous mathematical definition for the task. Figure \ref{fig:figure-1} is an illustration of the task.

After defining the task, we establish baselines for the benchmark and measure the collaboration among robots. In recent years, multi-agent reinforcement learning (MARL) has shown promising results in dealing with collaboration among multiple agents. For instance, the MARL algorithm based on value decomposition\cite{c30,c31,c32,c33,c34} measures the contribution of each individual agent to the team through credit assignment. The communication-based MARL algorithm\cite{c26,c27,c28,c29} explains the collaboration among robots by selecting the communication targets and contents among agents. And some of the MARL methods\cite{c18,c19,c25,c21} use the framework of "centralized learning, decentralized execution" (CTDE) to implicitly integrate the collaboration in agent teams into the decentralized policy of each agent. Therefore, we design several baseline methods based on the state-of-the-art MARL algorithms. Additionally, we introduce learnable target robots in the scenario, where all robots in the scenario are learnable, called the fully competitive game. In such a co-evolutionary process, the target robot and captor robot can mutually promote and improve each other.

In summary, our contributions include:

(1) We define a new benchmark to evaluate multi-robot collaboration in multi-robot target trapping task and propose corresponding task evaluation indicators;

(2) For the target trapping task, we build a 2D simulation environment named Target Trapping Environment (T2E) based on the real-world obstacle model;

(3) For this benchmark, we design several MARL algorithms as the baseline, conduct a series of experiments on the evaluation indicators, and provide insights into regimes of multi-robot collaboration.

\section{Related Work}

In this section, we will review the methods for multi-robot hunting task in recent years, which can be mainly divided into three categories: optimization-based, heuristic-based, and learning-based methods. 

Methods based on optimization theory usually abstract the problem into the form of objective functions with limited conditions, trying to theoretically guarantee the optimality of the solution\cite{c12,c1,c13,c11,c15,c2}. For example, Huang \emph{et al.} \cite{c12} utilized the Voronoi partition to transform the pursuit problem into a solution to the change of the partition area on a simple plane. Zhou \emph{et al.} \cite{c1} used the FMM (Fast Marching Method) to extend the Voronoi partition method to the environment with obstacles. Furthermore, Tian \emph{et al.}\cite{c15} proposed the free space division methods with the safe region under regular-shape obstacles, which guaranteed the robot without colliding with obstacles. On the other side, Chen \emph{et al.}\cite{c2} extended the fishing game model to the situation of multiple players and deduced the initial condition setting and robot movement strategy that can guarantee the success of the task under ideal geometric conditions. Most methods based on optimization theory consider perfect motion states or scenes with unnatural regular obstacles. Although they can theoretically obtain or approach the optimal solution, they usually suffer from the disadvantages of high computational complexity and poor generalization. 

Heuristic methods are usually intuitive or empirical, and the common ones are based on simulated virtual force or biologically inspired. Angelani \emph{et al.} \cite{c3} proposed the statement on the basis of the Vicsek model\cite{c5} that simulated forces can be used to characterize the predation phenomena between predator groups and prey groups. Janosov \emph{et al.} \cite{c4} considered that hunters and prey will be subject to virtual forces from teammates, obstacles, and opponents in complex scenarios with inertia, time delay, and noise to drive complex strategies. Although heuristic methods have low computational complexity, they are prone to fall into local optima and cannot guarantee the optimality of the solution. 

Learning-based methods usually use Reinforcement Learning (RL) to approach the optimal solution and the model usually has low computational complexity and well scalability\cite{c6,c7,c8,c9,c10}. For example, H{\"u}ttenrauch \emph{et al.} \cite{c6} proposed to use sampling and mean embedding to aggregate neighbors' information, which improved the scalability of the discrete strategy in the pursuit-evasion task. Souza \emph{et al.} \cite{c7} designed team rewards with formation score and used the combination of TD3 (Twin Delayed Deep Deterministic Policy Gradient) algorithm with curriculum learning to improve the training performance. Zhang \emph{et al.}\cite{c8} considered the environment with obstacles and combined the artificial potential field method with reinforcement learning.

\section{T2E: the Target Trapping Environment}

This section will give a detailed introduction to the multi-robot target trapping task. Similar to the multi-robot hunting task, there are also two types of robots that fight against with each other, which are called captor robots and target robots respectively. However, unlike the hunting task, the multi-robot target trapping task requires captor robots to trap the target robot until it cannot move. To this end, we refer to \cite{c1} and define a concept called \emph{Absolutely Safe Zone} (ASZ) to establish a mathematical expression for the trapping process. After establishing a mathematical description, the entire task is constructed as a Markov Decision Process and solved using MARL baseline algorithms. We will also explain the definitions of task-related actions, states, and rewards in this section.

\subsection{Problem Formulation} 
In the multi-robot target trapping task, the captor robot should use obstacles to restrict the target robot as much as possible, and the target robot needs to escape the containment of the target robot. Therefore, we define \emph{Absolutely Safe Zone} to describe the restricted degree of the target robot:

\textbf{Definition 1}: Suppose there is a robot at position $x(t)\in \mathcal{S}$ at time $t$, where $\mathcal{S}$ is a two-dimensional free space with obstacles. For any other point $y\in\mathcal{S}$, we define an \emph{arrival function} $f$ representing the minimum time required for the robot to move from the current position to point $y$:
\begin{equation}
f(x(t),y)=min\{t'-t|x(t')=y,y\in\mathcal{S},\forall t'>t\}
\end{equation}
In this paper, the FMM\cite{c17} algorithm is used to calculate the shortest path length for robots from the current position to a certain point on the map. Concretely, the arrival function of the robot can be expressed as the ratio of the shortest path length and speed of the robot.

\textbf{Definition 2}: Given $K$ captor robots, one target robot and a two-dimensional free space $\mathcal{S}$ with obstacles. The positions of the captor robot $i$ and the target robot at time $t$ are denoted as $x^p_i(t)$ and $x^e(t)$ respectively. And the arrival functions of the two are $f_p(\cdot)$ and $f_e(\cdot)$ respectively, then the \emph{Absolutely Safe Zone} of the target robot at time $t$ can be expressed as $\mathcal{S}_a(t)$:
\begin{equation}
\begin{split}
\mathcal{S}_a(t) = \{y|f_p(x^p_i(t),y)>f_e(x^e(t),y),\\
y\in\mathcal{S},\forall i\in[0,K]\} 
\end{split}
\end{equation}
$\mathcal{S}_a(t)$ represents the position where the target can reach safely without being disturbed by the captor, so it is called the Absolutely Safe Zone. The area of $\mathcal{S}_a(t)$ indicates the degree of restriction of the target robot. The smaller the area, the more dangerous the target robot is. In the multi-robot target trapping task, the goal of the captor robot is to reduce the Absolutely Safe Zone of the target robot, while the goal of the target robot is to expand its Absolutely Safe Zone as much as possible. If the area of the Absolutely Safe Zone is smaller than a certain threshold so that the target robot can no longer move, then the captor robots are considered to have captured the target robot. Therefore, a complete definition of the multi-robot target trapping task is given as follows:

\textbf{Definition 3}: Given $K$ captor robots, one target robot and a two-dimensional free space $\mathcal{S}$ with obstacles. The initial positions of the captor robot $i$ and the target robot are $x^p_i(0)$ and $x^e(0)$ respectively. The initial Absolutely Safe Zone can be denoted as $\mathcal{S}_a(0)$. And the control signals of the captor robot $i$ and the target robot at time $t$ are $ a^i_p (t)\in\mathcal{A} $ and $a_e(t)\in\mathcal{A} $ respectively where $\mathcal{A}$ is the action space. With a time budget of $T$, the multi-robot target trapping task can be formulated as:

\begin{equation}
\min_{{\{a_i^p(t)\}}_{i=1}^{K}}|\mathcal{S}_a(T)-\mathcal{S}_a(0)| 
\end{equation}
$$
\begin{aligned}
s.t.\  
& x^p_i(t+1)=g_p(x^p_i(t),a_i^p(t)) \ \forall t\in [0,T] \\
& x^e(t+1)=g_e(x^e(t),a^e(t)) \ \forall t\in [0,T] \\
& x^p_i(t),x^e(t) \in \mathcal{S} \ \forall t\in [0,T] \\
\end{aligned}
$$
where $g_p(\cdot)$ and $g_e(\cdot)$ are the motion equations of captor robots and the target robot, respectively. Note that this is an optimization goal from the perspective of the captor robot, so the goal is to minimize the size of the Absolutely Safe Zone within the time budget $T$.

\begin{figure}
    \centering
    \includegraphics[width=8.65cm,height=4.865cm]{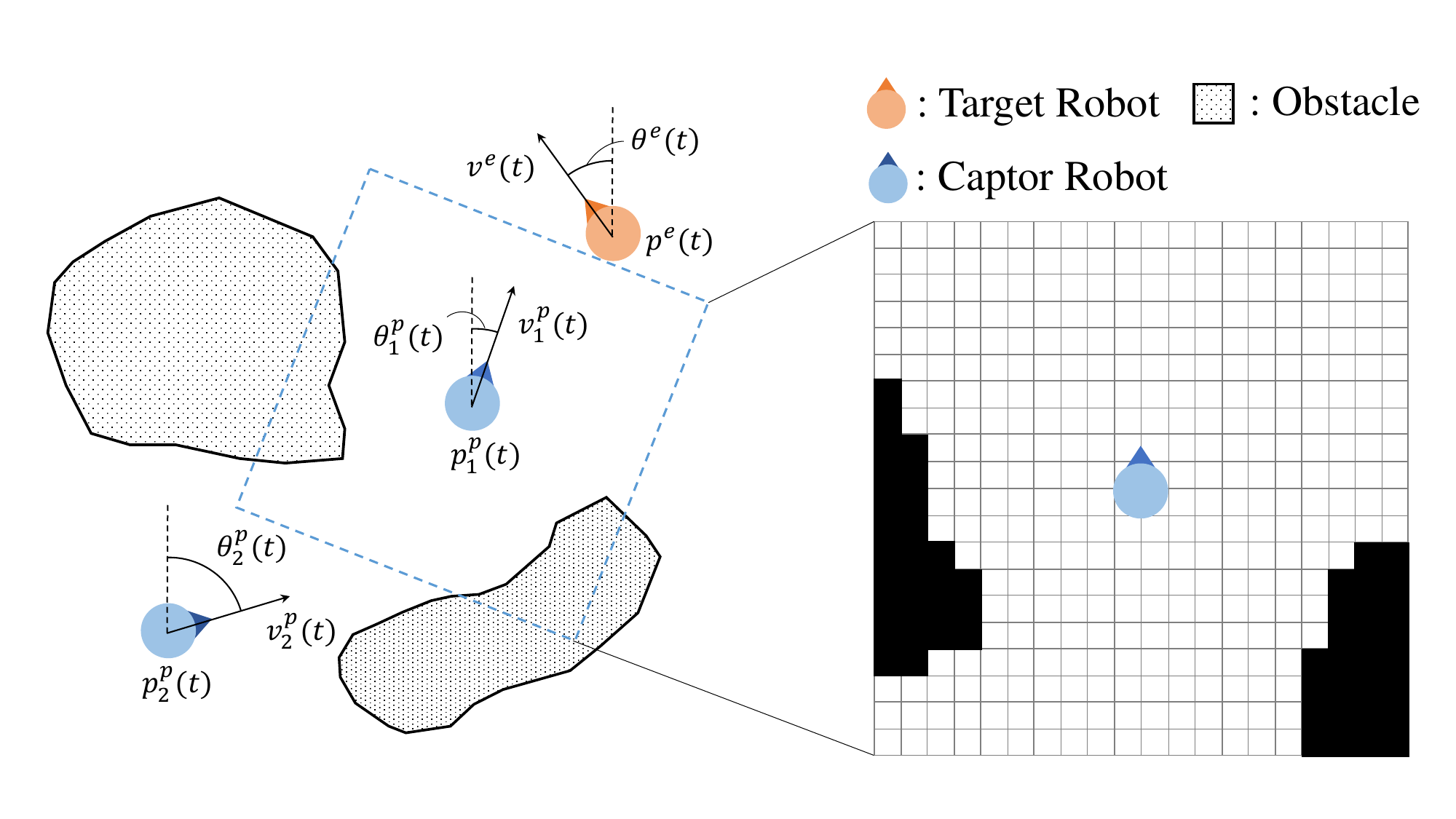}
    \caption{An illustration of Observation space. The captor robot can observe the entire internal state of teammates, but cannot observe the speed of the target robot. The target robot can observe the entire internal state of the captor robot. In addition, each robot can observe the obstacle information at a certain distance around itself.}
    \label{fig:figure0}
    \vspace{-0.3cm}
    
\end{figure}

\subsection{Action Space}

When designing the action space, we consider the simulation of a ground mobile robot in the real world. So the action space is set as discrete $5$ values on 2D plane: [ \emph{forward}, \emph{turnleft}, \emph{turnright}, \emph{stop}, \emph{backward} ]. \emph{turnleft} and \emph{turnright} represent a 30-degree left turn and a 30-degree right turn respectively. In the motion actuator, each action is translated into an action force $\bm{f_a}\in \mathbb{R}^2$ in the corresponding direction, driving the robot to move or stop. 

In addition, the artificial potential field method is adopted for obstacle avoidance. Once the distance between the robot and the obstacle is less than the danger threshold, the robot will be forced to turn or stop by the virtual repulsion force from the obstacle. Since the obstacles in the scene are irregular, we simplify the repulsion to come from $M$ directions around the robot, which can be expressed as follows:
\begin{equation}
\bm{f_v} = \sum_{m=1}^M{(-log(\frac{d_{\theta_m}}{k\cdot r})+1)\cdot \mathbb{I}(d_{\theta_m} < k\cdot r)\cdot \bm{u_{\theta_m}}}    
\end{equation}
where $\bm{u_{\theta_m}}$ is the unit vector in the direction of $\theta_m$ and $d_{\theta_m}$ is the distance between the robot and the nearest obstacle along the $\theta_m$ direction. $r$ is the size of the robots. $k$ is an adjustable parameter used to control the danger threshold. The robot finally moves under the combined force of $\bm{f_a}$ and $\bm{f_v}$. Note that although the value of the combined force is arbitrary, the acceleration of the robot will not exceed the specified maximum value.


\subsection{Observation Space}
\subsubsection{agent-related observation} In this paper, we represent the internal state of the robot as a combination of velocity, position, and orientation. Let $s_i^{p1}(t)=[v_i^p(t),p_i^p(t),\theta_i^p(t)]$, $s^{e1}(t)=[v^e(t),p^e(t),\theta^e(t)]$ denote the internal state of captor robot $i$ and target robot respectively. It is assumed that the captor robot can obtain the internal state of teammates through communication, but can only observe the position and orientation of the target robot. The observation of the captor robot $i$ on the teammate $k$ is denoted as $s_{ik}^{p1}(t)=[v_i^p(t),p_i^p(t)-p_k^p(t),\theta_i^p(t)-\theta_k^p(t)]$. And the observation of the captor robot $i$ on the target robot is $s_i^{e1}(t) = [p_i^p(t)-p^e(t),\theta_i^p(t)-\theta^e(t)]$. Therefore, the agent-related observation of captor robot $i$ can be denoted as $o^{p1}_i(t)=[s_i^{p1}(t),s_{i1}^{p1}(t),s_{i2}^{p1}(t),...,s_{iK}^{p1}(t), s_{i}^{e1}(t)]$. The agent-related observation of target robot can be denoted as $o^{e1}(t)=[s^{e1}(t),s_{e1}^{p1}(t),s_{e2}^{p1}(t),...,s_{eK}^{p1}(t)]$ where $s_{ei}^{p1}(t) = [v_i^p(t), p^e(t)-p_i^p(t),\theta^e(t)-\theta_i^p(t)]$ is the agent-related observation of target robot on captor robot $i$.

\subsubsection{obstacle-related observation} In order to enable robots to perceive the surrounding environment and utilize obstacles to contain or escape while moving, the obstacle-related observation is incorporated to each robot on the basis of agent-related observations. As shown in Fig. \ref{fig:figure0}, the local map is discretized at a certain resolution. Then the obstacle-related observation of each robot can be expressed as a mask matrix $\bm{G}\in\mathbb{R}^{d\times d}$ centered on the robot and following the rotation of the robot. Each binary element in the matrix represents whether its corresponding position is obstacle.

\subsection{Rewards}
Although the intuitive evaluation metric for the final task is the size of the Absolutely Safe Zone, we manually design denser rewards to aid training. The reward at each step is divided into two parts: competition reward $R^c_i(t)$ and private reward $R^s_i(t)$. For captor robot $i$, the competition reward $R^c_i(t)$ is:
\begin{equation}
R^c_i(t) = 
\begin{cases} 
k_1\cdot(d_i^e(t)-d_i^e(t-1)),  & \mbox{each } t \\
k_2\cdot\mathbb{I}(d_i^e(t)<d_{cs}),  & \mbox{each } t \\
k_3\cdot\mathbb{I}(|\mathcal{S}_a(t)|<f_{thre}) & \mbox{each } t \\
\end{cases} 
\end{equation}
where $d_i^e(t)$ is the distance from the target robot to captor robot $i$ at time $t$. And $d_{cs}$ is the collision threshold between robots. $f_{thre}$ is the minimum area value of the Absolutely Safe Zone. The target robot will be considered to be captured and the task will terminate once the area of Absolutely Safe Zone falls below $f_{thre}$. The private reward is designed to encourage robots to explore and avoid collisions:
\begin{equation}
R^s_i(t) = 
\begin{cases} 
-0.4,  & \mbox{each } t \\
-1,  & \mbox{if } |\bm{f_v}|>0 \\
\end{cases} 
\end{equation}

In the fully-competitive environment, the competitive reward of the target robot and the reward of the captor robot are zero-sum, while the private reward is calculated in the same way.

\begin{figure}
    \centering
    \includegraphics[width=8.65cm,height=4.865cm]{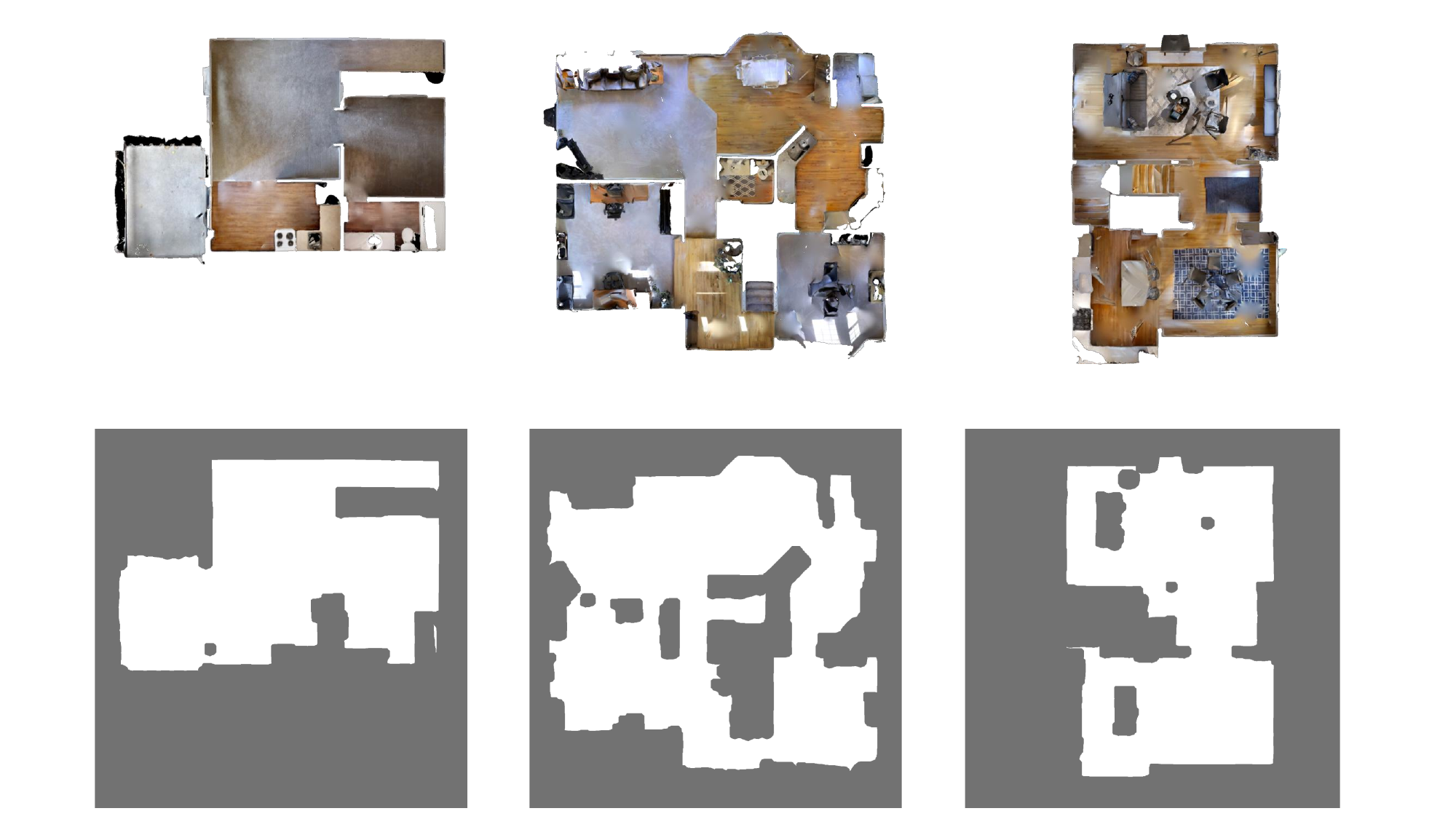}
    \caption{The examples of maps in the medium and large levels selected from the Gibson dataset. The lower row is the traversable map of the scene, and the upper row is the corresponding bird-view map.}
    \label{fig:figure1}
    \vspace{-0.3cm}
\end{figure}

\section{Experimental Setup}
\subsection{Simulation Environments}
To make the training scene close to the real world, we select $26$ top-down maps for training from the Gibson dataset\cite{c16}, which is widely used in robot navigation tasks. The average size of the traversable areas of the selected maps is $42.51 m^2$. During training, the map of each episode is randomly selected from these maps. When testing, the maps are divided into three sets of \emph{small}, \emph{medium}, and \emph{large} according to the size of the traversable area. The traversable area of the map in the \emph{small} level is smaller than $40 m^2$, and that in the \emph{medium} level is between $40m^2$ and $60m^2$. Those with a passable area greater than $60 m^2$ are classified as \emph{large}. A part of the training map is shown in Fig. \ref{fig:figure1}.

The robot's initial positions are randomly generated during both training and testing. The maximum initial distance between captor robots is $4m$, and that between target robots and captor robots is $10m$. Moreover, the maximum linear acceleration of the predator and target robots are $3m/s^2$ and $4m/s^2$ respectively. The radius of all robots is $0.2m$, and the maximum turning angle within one timestep is $\pi/6$. Note that calculating the size of the ASZ using the FMM algorithm during training consumes a lot of computing resources. So in practice, the target robot is considered to be captured if it will collide no matter which action it takes.

\subsection{MARL Methods}
\begin{figure}
    \centering
    \includegraphics[width=8.65cm,height=3.59cm]{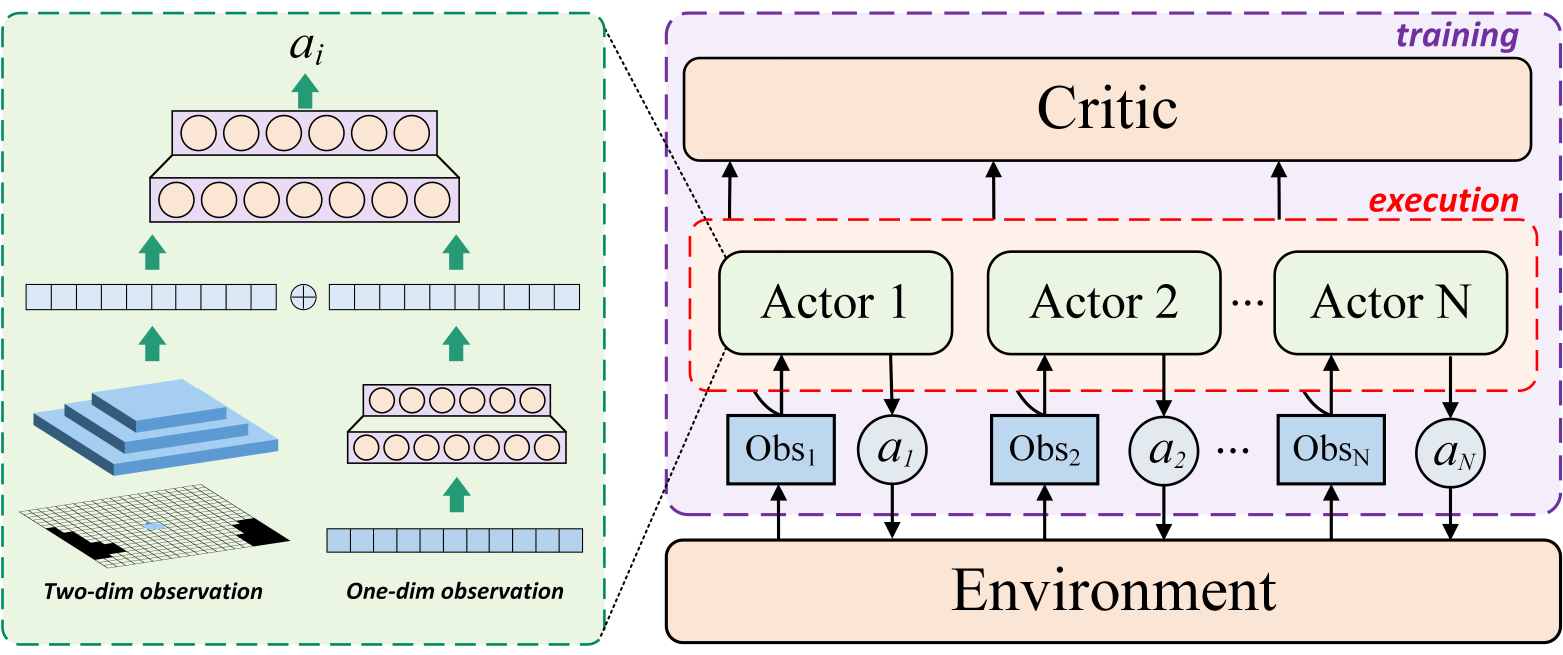}
    \caption{CTDE training framework and policy network (taking MAPPO as an example). Homogeneous agents share a centralized critic and decentralized actor during training. The policy network utilizes CNN and MLP to encode the observations respectively. The feature vectors will be concatenated together and fed into subsequent networks to output the current action.}
    \label{fig:figure2}
    \vspace{-0.3cm}
\end{figure}

Our goal is to find and verify the collaboration between robots through this benchmark. Therefore, we design four baselines based on algorithms currently widely used in the field of MARL: MADDPG\cite{c18}, MAAC\cite{c19}, IPPO\cite{c20}, and MAPPO\cite{c21}. For these baselines (except IPPO), we adopt the CTDE paradigm, which implicitly encodes the collaboration between agents into individual policies. Under this framework, homogeneous agents in the same team will share a centralized value network and a decentralized policy network. The details of the MARL framework and policy networks are shown in Fig. \ref{fig:figure2}.

In the implementation, the policy network consists of an encoding module and a decision module connected in series. The encoding module is composed of Multi-layer Perceptron (MLP) networks and Convolutional Neural Networks (CNN), which are used to encode the agent-related and obstacle-related observations respectively. Then the agent-related and obstacle-related features are concatenated together and sent to the subsequent decision module. The decision module composed of MLP and Recurrent Neural Networks (RNN) will output the action of the robot. As for the value network, a similar structure is adopted, where the observations (or observation-action pairs) of multiple robots are encoded separately by the encoding module. Then the features of multiple robots will be concatenated together, and the centralized state value (or action value) will be output through the subsequent network. Note that IPPO uses a discretized value network whose input is only the observation of the single robot. And for the same parameter quantity, the attention module in MAAC is replaced with MLP.

\begin{table*}[]
\renewcommand\arraystretch{1.2}
\caption{Quantitative results of baselines with different speed ratios. The ratio of captor robots to target robots is $3:1$. }
\centering
\begin{tabular}{|cccccccccc|}
\hline
\multicolumn{10}{|c|}{$v_p:v_e = 1.0:1.0$, $n_p:n_e=3:1$} \\ \hline
\multicolumn{1}{|c|}{Level} & \multicolumn{3}{c|}{Small} & \multicolumn{3}{c|}{Medium} & \multicolumn{3}{c|}{Large} \\ \hline
\multicolumn{1}{|c|}{Methods} & Time (step) & SR (\%) & \multicolumn{1}{c|}{Path Len (m)} & Time (step) & SR (\%) & \multicolumn{1}{c|}{Path Len (m)} & Time (step) & SR (\%) & Path Len (m) \\ \hline
\multicolumn{1}{|c|}{MADDPG} & 78.55 & 99.6 & \multicolumn{1}{c|}{7.410} & 109.34 & 99.8 & \multicolumn{1}{c|}{10.591} & 205.31 & 86.0 & 18.437 \\ \hline
\multicolumn{1}{|c|}{MAAC} & 68.02 & 100 & \multicolumn{1}{c|}{6.798} & 94.94 & 99.8 & \multicolumn{1}{c|}{9.066} & 167.47 & 91.2 & 15.125 \\ \hline
\multicolumn{1}{|c|}{MAPPO} & 100.71 & 99.8 & \multicolumn{1}{c|}{9.196} & 129.52 & 99.8 & \multicolumn{1}{c|}{12.261} & 185.44 & 94.8 & 17.289 \\ \hline
\multicolumn{1}{|c|}{IPPO} & 121.45 & 99.4 & \multicolumn{1}{c|}{11.249} & 182.63 & 95.8 & \multicolumn{1}{c|}{17.305} & 215.69 & 94.0 & 18.796 \\ \hline
\end{tabular}

\begin{tabular}{|cccccccccc|}
\hline
\multicolumn{10}{|c|}{$v_p:v_e = 1.0:1.2$, $n_p:n_e=3:1$} \\ \hline
\multicolumn{1}{|c|}{Level} & \multicolumn{3}{c|}{Small} & \multicolumn{3}{c|}{Medium} & \multicolumn{3}{c|}{Large} \\ \hline
\multicolumn{1}{|c|}{Methods} & Time (step) & SR (\%) & \multicolumn{1}{c|}{Path Len (m)} & Time (step) & SR (\%) & \multicolumn{1}{c|}{Path Len (m)} & Time (step) & SR (\%) & Path Len (m) \\ \hline
\multicolumn{1}{|c|}{MADDPG} & 174.08 & 91.4 & \multicolumn{1}{c|}{16.822} & 255.14 & 84.3 & \multicolumn{1}{c|}{24.765} & 342.58 & 60.8 & 32.178 \\ \hline
\multicolumn{1}{|c|}{MAAC} & 194.07 & 87.5 & \multicolumn{1}{c|}{19.880} & 291.24 & 73.6 & \multicolumn{1}{c|}{27.881} & 366.99 & 51.6 & 32.172 \\ \hline
\multicolumn{1}{|c|}{MAPPO} & 165.09 & 92.4 & \multicolumn{1}{c|}{16.572} & 243.93 & 80.6 & \multicolumn{1}{c|}{24.200} & 286.78 & 78.4 & 28.169 \\ \hline
\multicolumn{1}{|c|}{IPPO} & 222.14 & 84.8 & \multicolumn{1}{c|}{22.470} & 319.04 & 64.8 & \multicolumn{1}{c|}{29.214} & 323.13 & 68.4 & 31.944 \\ \hline
\end{tabular}

\begin{tabular}{|cccccccccc|}
\hline
\multicolumn{10}{|c|}{$v_p:v_e = 1.0:1.4$, $n_p:n_e=3:1$} \\ \hline
\multicolumn{1}{|c|}{Level} & \multicolumn{3}{c|}{Small} & \multicolumn{3}{c|}{Medium} & \multicolumn{3}{c|}{Large} \\ \hline
\multicolumn{1}{|c|}{Methods} & Time (step) & SR (\%) & \multicolumn{1}{c|}{Path Len (m)} & Time (step) & SR (\%) & \multicolumn{1}{c|}{Path Len (m)} & Time (step) & SR (\%) & Path Len (m) \\ \hline
\multicolumn{1}{|c|}{MADDPG} & 392.01 & 40.8 & \multicolumn{1}{c|}{37.395} & 431.75 & 27.4 & \multicolumn{1}{c|}{42.521} & 449.41 & 22.4 & 40.089 \\ \hline
\multicolumn{1}{|c|}{MAAC} & 358.89 & 49.8 & \multicolumn{1}{c|}{34.803} & 425.95 & 31.2 & \multicolumn{1}{c|}{40.822} & 421.57 & 32.8 & 40.083 \\ \hline
\multicolumn{1}{|c|}{MAPPO} & 306.84 & 64.2 & \multicolumn{1}{c|}{30.785} & 386.00 & 43.8 & \multicolumn{1}{c|}{37.662} & 397.94 & 43.1 & 37.123 \\ \hline
\multicolumn{1}{|c|}{IPPO} & 349.46 & 54.2 & \multicolumn{1}{c|}{30.180} & 389.92 & 41.3 & \multicolumn{1}{c|}{38.281} & 405.74 & 40.8 & 38.261 \\ \hline
\end{tabular}

\label{table:table2}
   \vspace{-0.2cm}
\end{table*}

\subsection{Evaluation Metrics}
To measure multi-robot target trapping performance, we evaluate the baselines in terms of efficiency and collaboration. Efficiency can be measured by the completion time of  the task, while collaboration can be measured by area change of ASZ. Therefore, evaluation metrics for captor robots of the multi-robot target trapping task can be summarized as follows:
\subsubsection{Time}Time refers to the cumulative time for captor robots to successfully capture the target robot, and it is a direct evaluation of efficiency.
\subsubsection{SR (Success Rate)}The success rate is the capture success rate of the captor robot under the limit of the maximum episode length. In this paper, the maximum episode length is set to $500$ during testing.
\subsubsection{Path Len (Path Length)}The path length refers to the average distance traveled by the captor robots in the task.
\subsubsection{SoA (Size of ASZ)}The area of ASZ represents the limit of multiple captor robots to the target robot. This metric is closely related to the relative positions between all captor robots and the target robot. Therefore, it can also be used to measure the collaboration between multiple captor robots.

\section{Experimental Result and Discussion}

\begin{table*}[]
\renewcommand\arraystretch{1.2}
\caption{Quantitative results of baselines with different speed ratios. The ratio of captor robots to target robots is $4:1$. }
\centering
\begin{tabular}{|cccccccccc|}
\hline
\multicolumn{10}{|c|}{$v_p:v_e = 1.0:1.0$, $n_p:n_e=4:1$} \\ \hline
\multicolumn{1}{|c|}{Level} & \multicolumn{3}{c|}{Small} & \multicolumn{3}{c|}{Medium} & \multicolumn{3}{c|}{Large} \\ \hline
\multicolumn{1}{|c|}{Methods} & Time (step) & SR (\%) & \multicolumn{1}{c|}{Path Len (m)} & Time (step) & SR (\%) & \multicolumn{1}{c|}{Path Len (m)} & Time (step) & SR (\%) & Path Len (m) \\ \hline
\multicolumn{1}{|c|}{MADDPG} & 119.33 & 98.6 & \multicolumn{1}{c|}{12.845} & 182.66 & 94.6 & \multicolumn{1}{c|}{16.966} & 290.12 & 73.2 & 27.114 \\ \hline
\multicolumn{1}{|c|}{MAAC} & 101.84 & 99.4 & \multicolumn{1}{c|}{9.849} & 172.67 & 97.6 & \multicolumn{1}{c|}{16.364} & 250.15 & 82.8 & 21.716 \\ \hline
\multicolumn{1}{|c|}{MAPPO} & 77.85 & 100 & \multicolumn{1}{c|}{7.086} & 107.56 & 99.8 & \multicolumn{1}{c|}{9.789} & 138.56 & 99.2 & 12.439 \\ \hline
\multicolumn{1}{|c|}{IPPO} & 82.98 & 100 & \multicolumn{1}{c|}{7.467} & 117.16 & 100 & \multicolumn{1}{c|}{11.114} & 136.78 & 99.6 & 13.008 \\ \hline
\end{tabular}

\begin{tabular}{|cccccccccc|}
\hline
\multicolumn{10}{|c|}{$v_p:v_e = 1.0:1.2$, $n_p:n_e=4:1$} \\ \hline
\multicolumn{1}{|c|}{Level} & \multicolumn{3}{c|}{Small} & \multicolumn{3}{c|}{Medium} & \multicolumn{3}{c|}{Large} \\ \hline
\multicolumn{1}{|c|}{Methods} & Time (step) & SR (\%) & \multicolumn{1}{c|}{Path Len (m)} & Time (step) & SR (\%) & \multicolumn{1}{c|}{Path Len (m)} & Time (step) & SR (\%) & Path Len (m) \\ \hline
\multicolumn{1}{|c|}{MADDPG} & 360.510 & 51.6 & \multicolumn{1}{c|}{32.786} & 420.558 & 31.8 & \multicolumn{1}{c|}{40.036} & 472.699 & 14.6 & 40.996 \\ \hline
\multicolumn{1}{|c|}{MAAC} & 184.686 & 89.8 & \multicolumn{1}{c|}{16.421} & 259.948 & 79.8 & \multicolumn{1}{c|}{24.748} & 325.130 & 62.4 & 29.216 \\ \hline
\multicolumn{1}{|c|}{MAPPO} & 98.308 & 99.0 & \multicolumn{1}{c|}{9.703} & 158.556 & 93.6 & \multicolumn{1}{c|}{15.36} & 193.316 & 93.8 & 18.127 \\ \hline
\multicolumn{1}{|c|}{IPPO} & 98.890 & 99.2 & \multicolumn{1}{c|}{9.455} & 137.244 & 97.8 & \multicolumn{1}{c|}{13.371} & 182.844 & 95.6 & 17.299 \\ \hline
\end{tabular}

\begin{tabular}{|cccccccccc|}
\hline
\multicolumn{10}{|c|}{$v_p:v_e = 1.0:1.4$, $n_p:n_e=4:1$} \\ \hline
\multicolumn{1}{|c|}{Level} & \multicolumn{3}{c|}{Small} & \multicolumn{3}{c|}{Medium} & \multicolumn{3}{c|}{Large} \\ \hline
\multicolumn{1}{|c|}{Methods} & Time (step) & SR (\%) & \multicolumn{1}{c|}{Path Len (m)} & Time (step) & SR (\%) & \multicolumn{1}{c|}{Path Len (m)} & Time (step) & SR (\%) & Path Len (m) \\ \hline
\multicolumn{1}{|c|}{MADDPG} & 409.44 & 30.6 & \multicolumn{1}{c|}{39.399} & 450.69 & 22.2 & \multicolumn{1}{c|}{43.645} & 474.04 & 11.0 & 44.251 \\ \hline
\multicolumn{1}{|c|}{MAAC} & 231.97 & 82.6 & \multicolumn{1}{c|}{22.201} & 330.91 & 59.6 & \multicolumn{1}{c|}{32.013} & 381.73 & 45.0 & 33.809 \\ \hline
\multicolumn{1}{|c|}{MAPPO} & 133.84 & 93.2 & \multicolumn{1}{c|}{13.027} & 199.70 & 87.2 & \multicolumn{1}{c|}{19.545} & 229.51 & 85.6 & 21.072 \\ \hline
\multicolumn{1}{|c|}{IPPO} & 150.87 & 94.0 & \multicolumn{1}{c|}{14.692} & 219.95 & 86.2 & \multicolumn{1}{c|}{21.502} & 237.94 & 87.6 & 22.780 \\ \hline
\end{tabular}
\label{table:table3}
\end{table*}

\subsection{Multi-robot Target Trapping Experiment}
To verify the performance of baseline methods in the multi-robot target trapping task, we conduct experiments under fully cooperative and fully competitive settings. In the fully cooperative setting, only the policy of the captor robot is learnable network, and the target robot adopts a rule-based strategy. In the fully competitive setting, policies of both captor robots and the target robot are learnable networks, and both sides use the latest updated policies in each episode. Their policies gradually co-evolve during the competition. And we conduct experiments under different speed comparisons and number comparisons of robots for both fully cooperative and fully competitive settings.

\begin{figure*}[!htb]
   \vspace{-1em}
  \centering
  \subfigure[$v_p:v_e = 1.0:1.0$]{\includegraphics[width=5.53cm,height=4.29cm]{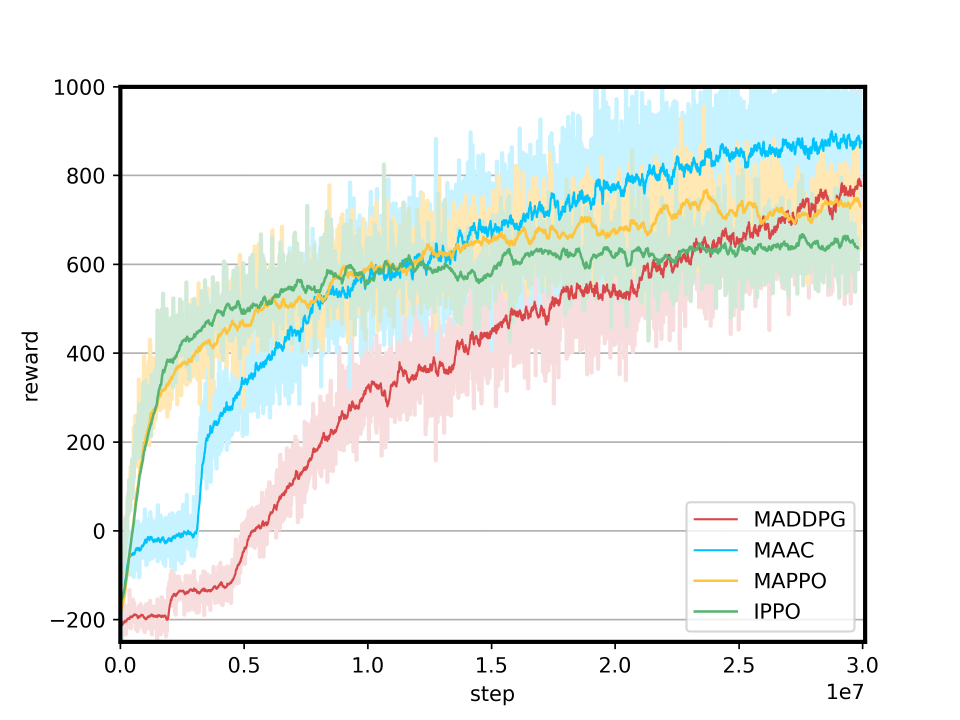}}
  \subfigure[$v_p:v_e = 1.0:1.2$]{\includegraphics[width=5.53cm,height=4.29cm]{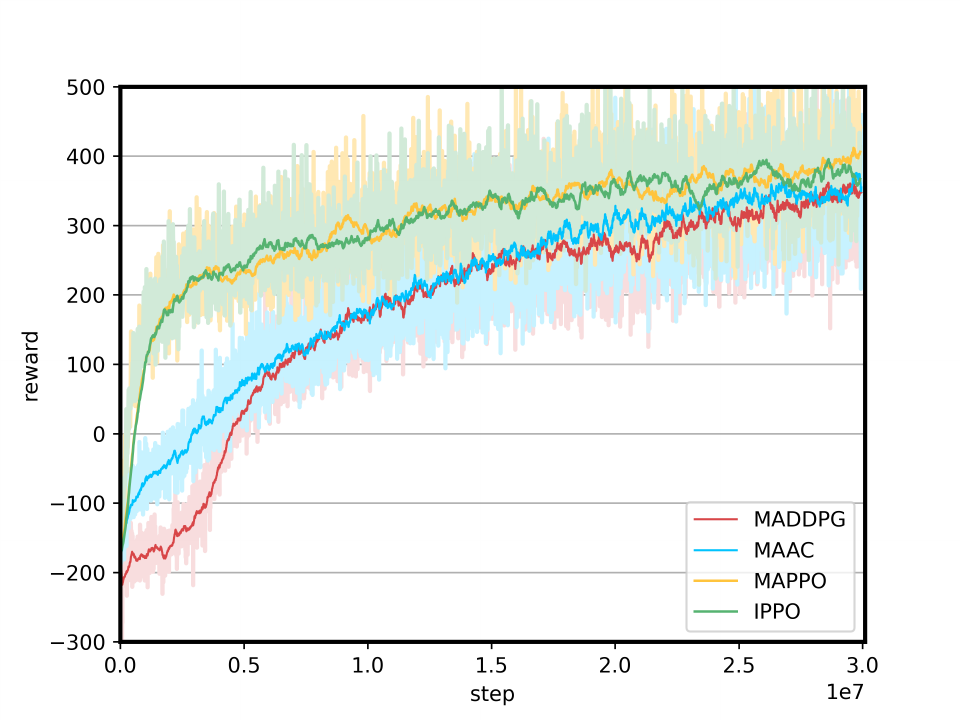}}
  \subfigure[$v_p:v_e = 1.0:1.4$]{\includegraphics[width=5.53cm,height=4.29cm]{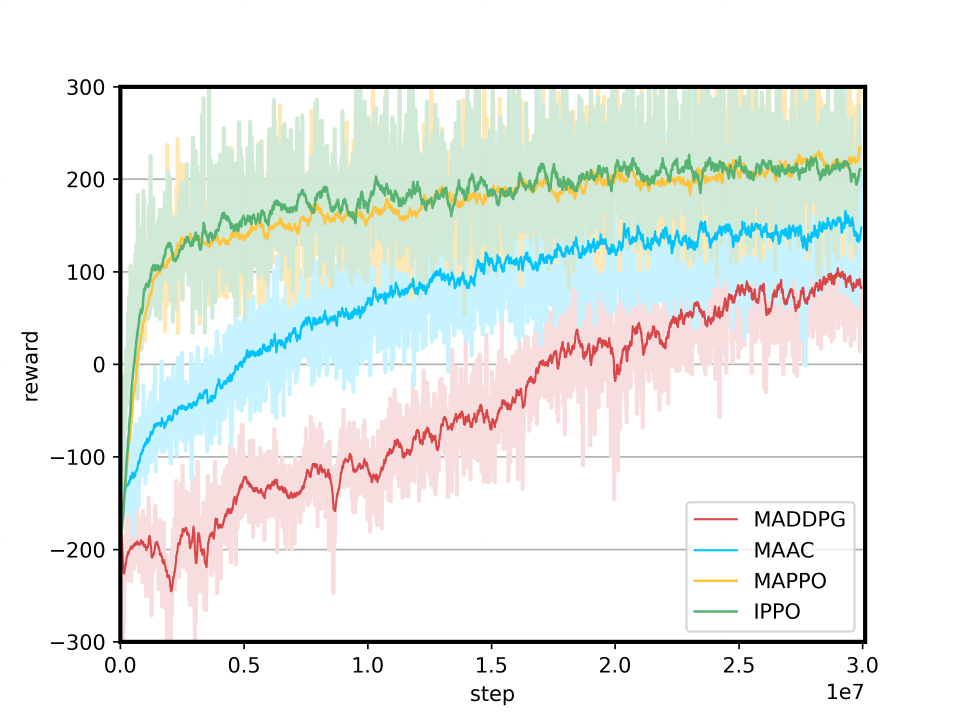}}
  \caption{Performance of the baselines in terms of training steps with $n_p:n_e=3:1$ under fully-cooperative setting.}
  \label{fig:figure3}
    \vspace{-0.3cm}
\end{figure*}
\begin{figure*}[!htb]
   \vspace{-1em}
  \centering
  \subfigure[$v_p:v_e = 1.0:1.0$]{\includegraphics[width=5.53cm,height=4.29cm]{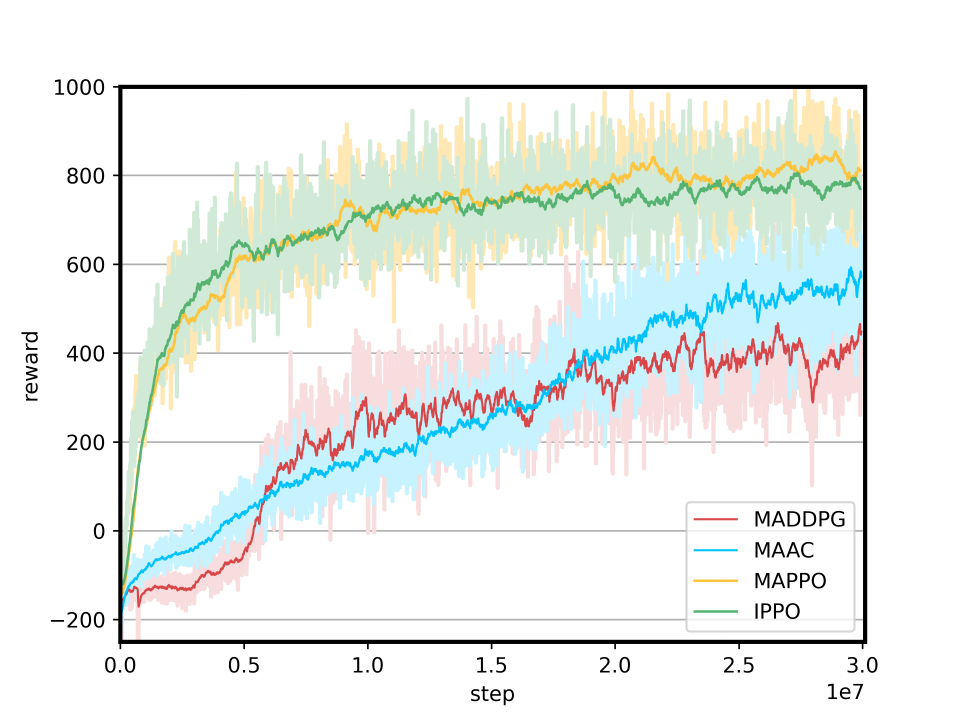}}
  \subfigure[$v_p:v_e = 1.0:1.2$]{\includegraphics[width=5.53cm,height=4.29cm]{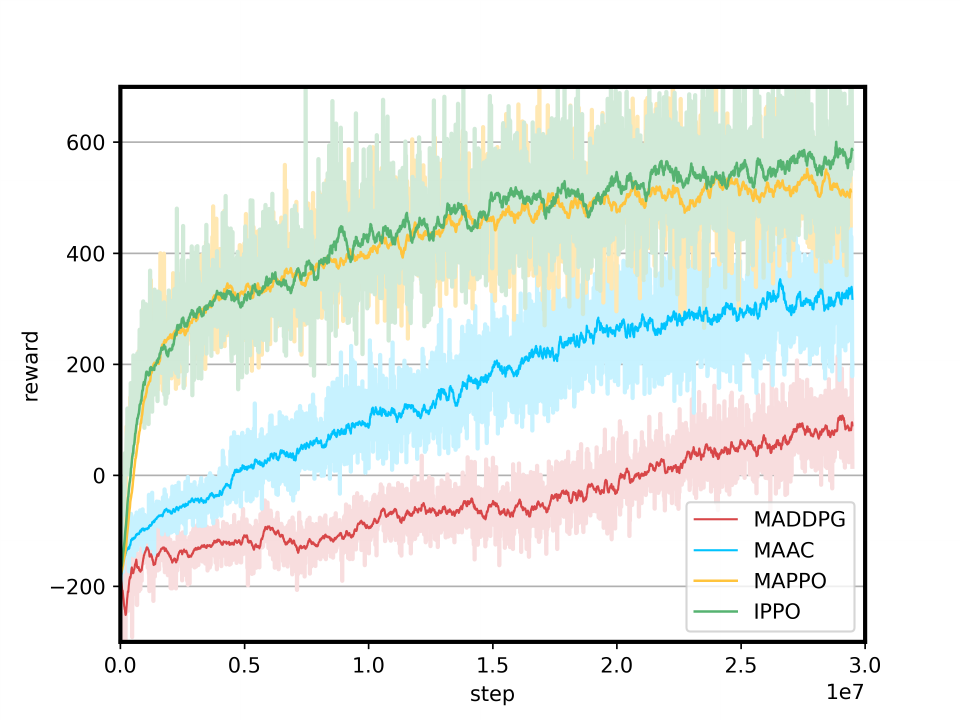}}
  \subfigure[$v_p:v_e = 1.0:1.4$]{\includegraphics[width=5.53cm,height=4.29cm]{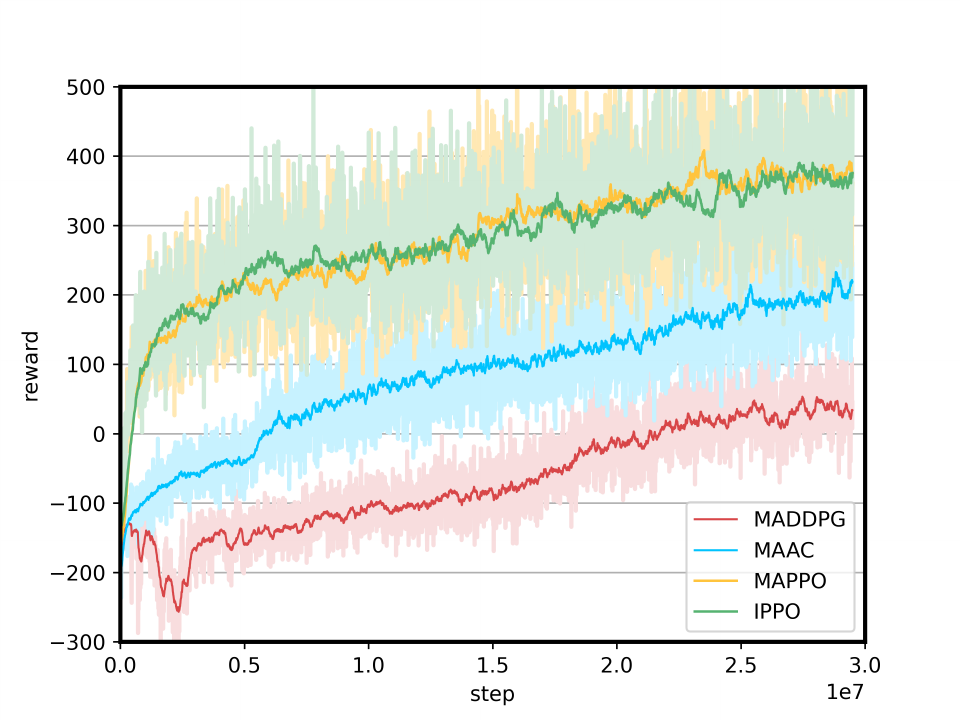}}
  \caption{Performance of the baselines in terms of training steps with $n_p:n_e=4:1$ under fully-cooperative setting.}
  \label{fig:figure4}
    \vspace{-0.3cm}
\end{figure*}

\begin{figure*}[!h]
  \centering
  \subfigure[Small level map]{\includegraphics[width=5.53cm,height=4.29cm]{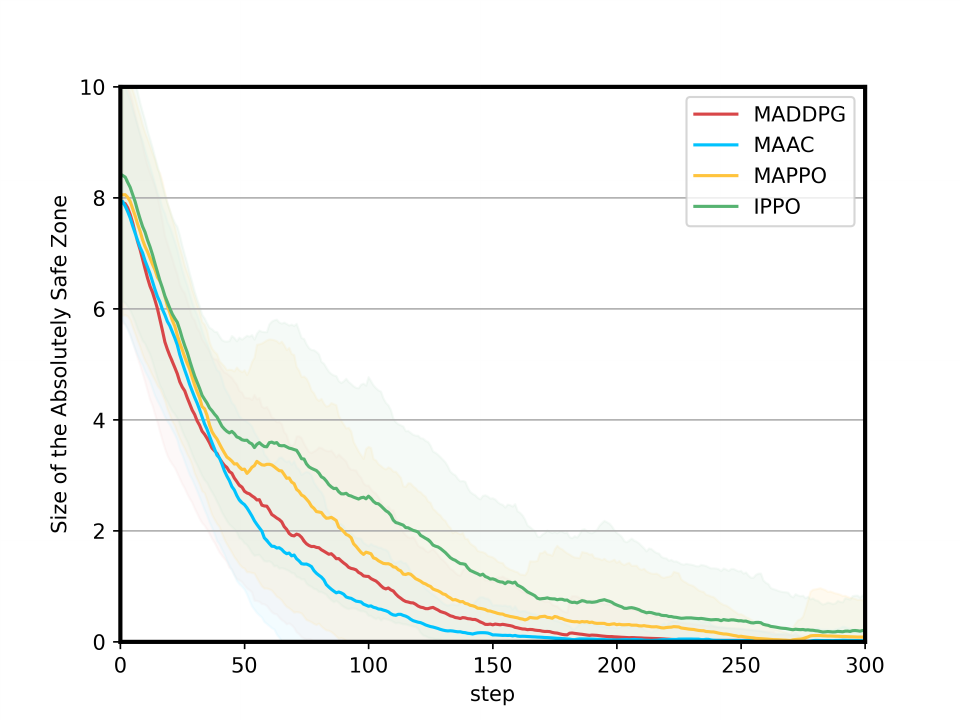}}
  \subfigure[Medium level map]{\includegraphics[width=5.53cm,height=4.29cm]{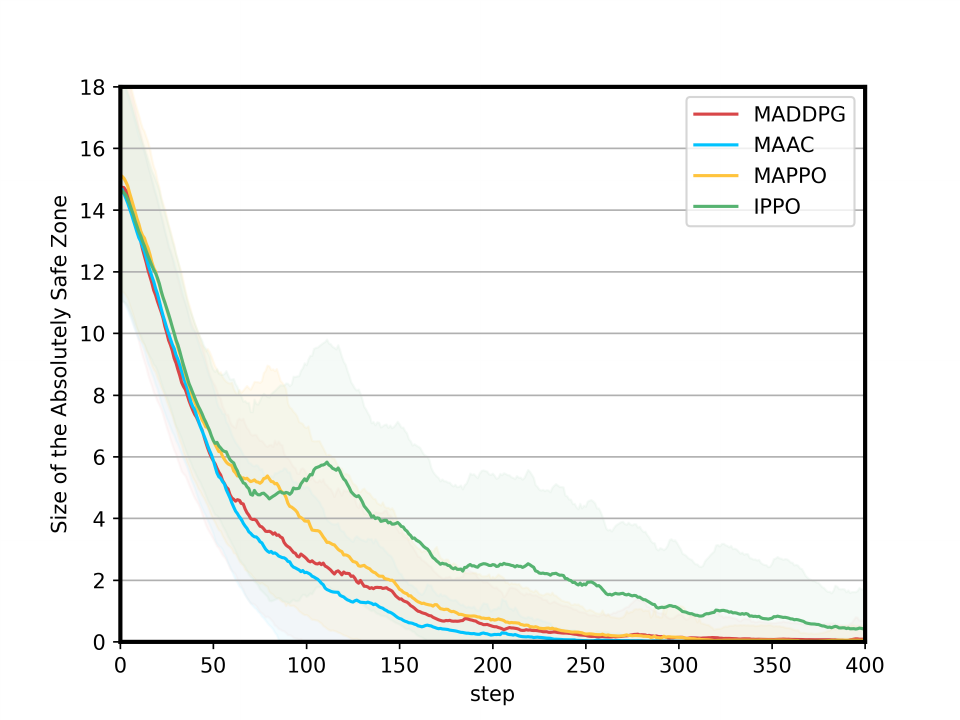}}
  \subfigure[Large level map]{\includegraphics[width=5.53cm,height=4.29cm]{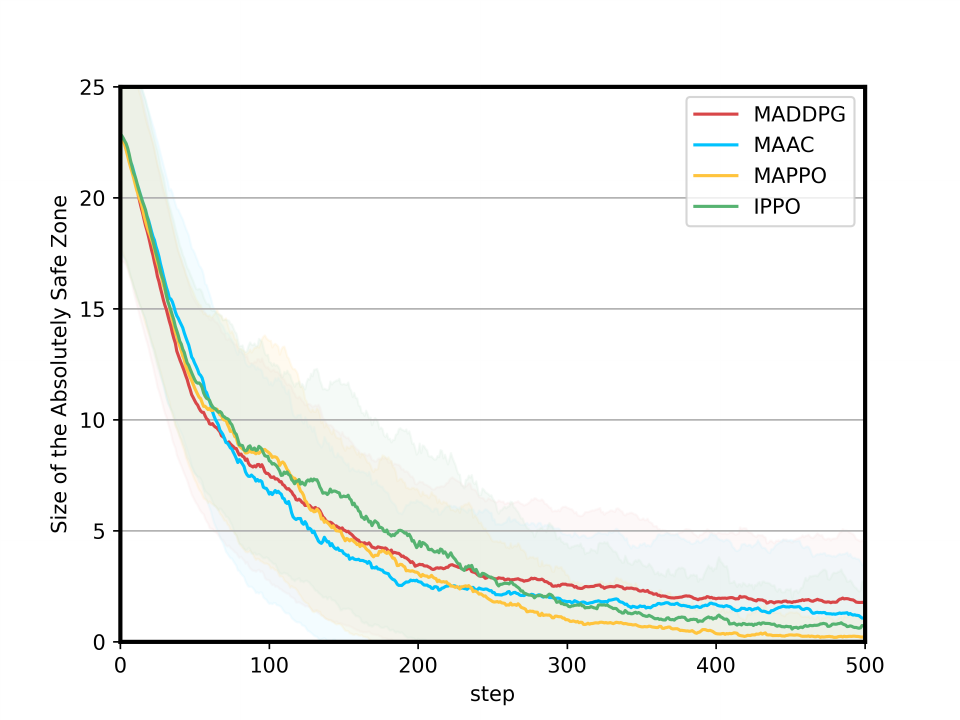}}
  \caption{Size of ASZ in maps of different levels in a test episode.}
  \label{fig:figure5}
      \vspace{-0.3cm}
\end{figure*}

\begin{figure}[!h]
   \vspace{-1em}
  \centering
  \includegraphics[width=6.636cm,height=5.268cm]{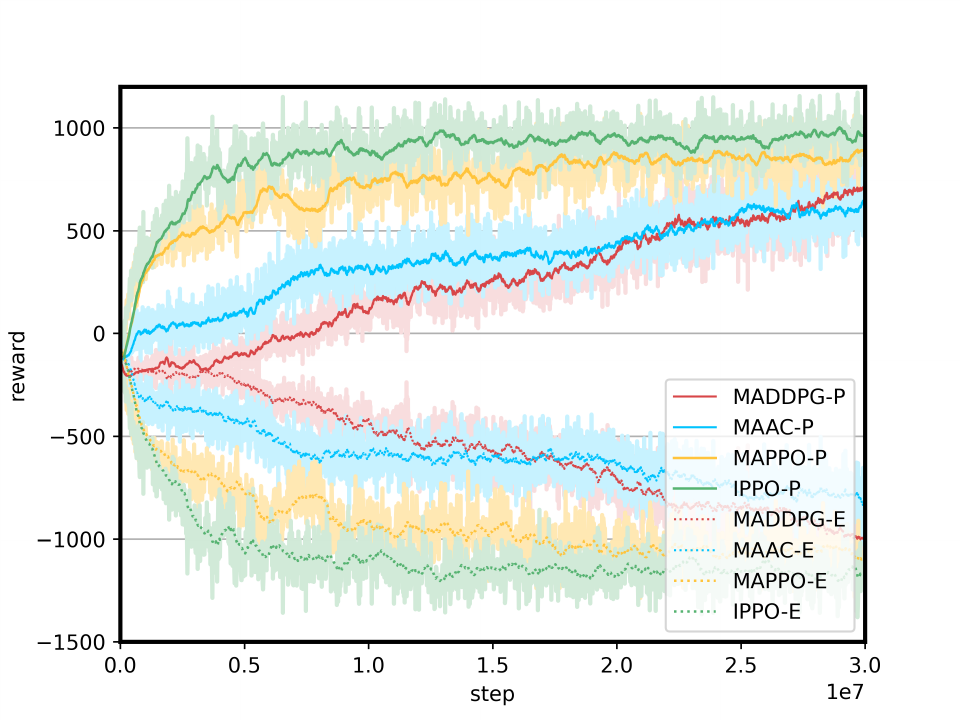}
  \caption{Performance of the baselines in terms of training steps with $n_p:n_e=3:1$ and $v_p:v_e=1.0:1.0$  under fully-competitive setting. MADDPG-P, MAAC-P, MAPPO-P, and IPPO-P denote the reward of captor robots, while MADDPG-E, MAAC-E, MAPPO-E, and IPPO-E denote the reward of the target robot.}
  \label{fig:figure6}
    \vspace{-0.1cm}
\end{figure}

\begin{table}[]
\renewcommand\arraystretch{1.1}
\centering
  \caption{Quantitative results of the trained captor robot under the fully-competitive setting.}
\begin{tabular}{|c|ccc|}
\hline
 & \multicolumn{3}{c|}{Rule-based target robot} \\ \hline
Methods & Time (step) & SR(\%) & Path Len (m) \\ \hline
MADDPG & 433.98 & 27.2 & 39.919 \\ \hline
MAAC & 139.26 & 96.6 & 13.779 \\ \hline
MAPPO & 159.02 & 95.8 & 15.709 \\ \hline
IPPO & 166.87 & 95.6 & 16.476 \\ \hline
\multicolumn{1}{|l|}{} & \multicolumn{3}{c|}{Network-policy target robot} \\ \hline
Methods & Time (step) & SR(\%) & Path Len (m) \\ \hline
MADDPG & 74.88 & 94.4 & 6.388 \\ \hline
MAAC & 112.59 & 98.0 & 10.603 \\ \hline
MAPPO & 46.08 & 99.8 & 4.429 \\ \hline
IPPO & 43.23 & 100 & 4.316 \\ \hline
\end{tabular}
\label{table:table5}
   \vspace{-0.2cm}
\end{table}

\subsubsection{Training in Fully-cooperative Game}
Under the fully-cooperative setting, the strategy of the target robot is based on the heuristic method in\cite{c4}, while the policies of captor robots are learnable networks. Captor robots need to fully cooperate to fight against rule-based target robots and continuously optimize their own policies during training. We conduct experiments with different ratios of captor robots to the target robot, including $3:1$ and $4:1$, and different speed ratios of captor robots to the target robot, including $1.0:1.0$, $1.0:1.2$, and $1.0:1.4$.

The training results of each experiment are shown in Figure \ref{fig:figure3}-\ref{fig:figure4}. And the depicted curves represent the post-smoothed curves, while the light-colored background corresponds to the pre-smoothed curves. The figures indicate that the task becomes progressively more challenging as the speed of the target robot increases, leading to a reduction in the rewards available to captor robots. MAAC and MADDPG can converge to a good reward score when the task setting is relatively simple. Their performance is worse than that of MAPPO and IPPO when the task difficulty increases.  And we test the proposed metrics after training, as shown in Table \ref{table:table2} and \ref{table:table3}. Note that the length of each episode is set to $200$ during training and $500$ during testing. The results presented in the tables indicate that MADDPG and MAAC exhibit inferior performance compared to MAPPO and IPPO, especially in the case of $n_p:n_e=4:1$. This disparity can be partially attributed to the increased level of coordination required as the number of robots increases. In our task setting, the robots' geometric models are designed to closely resemble those of ground robots in real-world scenarios, which places limitations on their motion abilities. As a result of their constrained omnidirectional mobility, the captor robots experience motion interference when they approach one another, thereby increasing the demands on their coordination. Additionally, it should be noted that both MADDPG and MAAC are off-policy algorithms, which are known to be more unstable during training when compared to on-policy algorithms. Hence, it is the combination of the high coordination demands inherent in the multi-robot target trapping task and the instability of the off-policy training process that leads to the observed performance differences.

In addition, we test the ASZ area change of the prey in the multi-robot target trapping task under the setting of $n_p:n_e=3:1$ and $v_p:v_e=1.0:1.0$. The experimental results in Figure \ref{fig:figure5} are the changes in the ASZ area over time in an episode. It can be seen that although the success rate of the IPPO algorithm is comparable to other algorithms, the decline rate of the ASZ area is generally slower than other algorithms. And in Table \ref{table:table2} and \ref{table:table3}, the average path length of the IPPO algorithm is longer than that of MAPPO, which means that IPPO needs to spend more travel distance and time in the target trapping. This shows that although the algorithm under the decentralized critic paradigm can learn to finally trap the prey, the cooperation in the trapping process is still not as good as that under the centralized critic.

\subsubsection{Training in Fully-competitive Game}
Under the fully-competitive setting, the policies of captor robots and the target robot are both learnable networks and participate in training together. During the training process, both two sides use the latest updated strategy in each episode. Since there is only one target robot, the learning framework of the target robot is a single-agent special case corresponding to the MARL framework. We conduct experiments under the setting that the number ratio of captor robots and the target robot is $3:1$, and the speed ratio is $1.0:1.0$.

The training results under the fully-competitive setting are shown in Figure \ref{fig:figure6}. Since the competitive rewards of the captor robot and the target robot are zero-sum, the rewards exhibit some symmetry. Furthermore, we observe that the rewards of captor robots converge to higher levels while those of the target robot converge to lower levels. To further verify the decision-making ability learned by the target robot under the fully-competitive setting, we confront learned captor robots with the learned target robot and the rule-based prey respectively. The experimental results are shown in Table \ref{table:table5}. From the table, it can be found that the success rate of the learned captor robot against the rule-based target robot is slightly lower than that of the learned target robot. There is even a large deviation between the two for MADDPG. This may be because the prey fails to converge during the evolution process of joint learning, which in turn causes the captor robot to overfit this failed target robot. Due to the lack of a stronger opponent, the captor robot itself cannot obtain stronger capabilities. 

\subsection{Discussion}
Based on the above results and analysis, we discuss the future research direction for the multi-robot target trapping task and provide some insights:

(1) An important finding is that increasing the number of robots may not necessarily make the task simpler, but instead imposes higher demands on collaboration. This is because in the real world, denser robots imply more complex planning conflicts under physical property constrain. This point can be illustrated by the convergence of MADDPG.

(2) Robots can take obstacles and boundary information into consideration, and use obstacles to cause trouble for their opponents. For example, the target robot can use a narrow passage to escape from the pursuing captor robot, while the captor robot can use the edge of the map to trap the target robot. How to extract and represent obstacle information of the environment in a more targeted manner is a direction worth studying in the future.

(3) It is possible to design a learning curriculum that can promote mutual improvement between the two opposing robots in a competition. In our experiment, the prey robot's convergence to a lower level results in weakness for both the prey and captor robots. However, with a targeted course design, both parties in the game should be able to take turns improving their abilities, thereby forcing their opponents to acquire stronger decision-making capabilities.

\section{Conclusion}
In this paper, we propose a new benchmark for multi-robot collaboration called the multi-robot target trapping task. After defining the task, we establish a 2D simulation environment and design several MARL algorithms as baselines, which we have made available as open-source code. And we also introduce corresponding evaluation metrics and conduct extensive experiments to evaluate the task. Finally, we analyze and discuss the experimental results, providing insights for future work on the multi-robot target trapping task. In future work, we will introduce more complex game frameworks to conduct further research on the common progress of robots in the training. In addition, how to effectively and interpretably encode obstacle information in the environment is also one of our research goals in future.


\bibliographystyle{Bibliography/IEEEtranTIE}
\bibliography{BIB_xx-TIE-xxxx}\ 

\begin{thebibliography}{10}
\providecommand{\url}[1]{#1}
\csname url@samestyle\endcsname
\providecommand{\newblock}{\relax}
\providecommand{\bibinfo}[2]{#2}
\providecommand{\BIBentrySTDinterwordspacing}{\spaceskip=0pt\relax}
\providecommand{\BIBentryALTinterwordstretchfactor}{4}
\providecommand{\BIBentryALTinterwordspacing}{\spaceskip=\fontdimen2\font plus
\BIBentryALTinterwordstretchfactor\fontdimen3\font minus
  \fontdimen4\font\relax}
\providecommand{\BIBforeignlanguage}[2]{{%
\expandafter\ifx\csname l@#1\endcsname\relax
\typeout{** WARNING: IEEEtran.bst: No hyphenation pattern has been}%
\typeout{** loaded for the language `#1'. Using the pattern for}%
\typeout{** the default language instead.}%
\else
\language=\csname l@#1\endcsname
\fi
#2}}
\providecommand{\BIBdecl}{\relax}
\BIBdecl

\bibitem{c38}
R.~Isaacs, \emph{Differential games: a mathematical theory with applications to
  warfare and pursuit, control and optimization}.\hskip 1em plus 0.5em minus
  0.4em\relax Courier Corporation, 1999.

\bibitem{c39}
M.~Pachter, ``Simple-motion pursuit-evasion in the half plane,''
  \emph{Computers \& Mathematics with Applications}, vol.~13, no. 1-3, pp.
  69--82, 1987.

\bibitem{c40}
T.~Ba{\c{s}}ar and G.~J. Olsder, \emph{Dynamic noncooperative game
  theory}.\hskip 1em plus 0.5em minus 0.4em\relax SIAM, 1998.

\bibitem{c36}
T.~Gokkul~Nath, P.~Sudheesh, and M.~Jayakumar, ``Tracking inbound enemy missile
  for interception from target aircraft using extended kalman filter,'' in
  \emph{Security in Computing and Communications: 4th International Symposium,
  SSCC 2016, Jaipur, India, September 21-24, 2016, Proceedings 4}, pp.
  269--279.\hskip 1em plus 0.5em minus 0.4em\relax Springer, 2016.

\bibitem{c37}
J.~Rao, X.~Xu, H.~Bian, J.~Chen, Y.~Wang, J.~Lei, W.~Giernacki, and M.~Liu, ``A
  modified random network distillation algorithm and its application in usvs
  naval battle simulation,'' \emph{Ocean Engineering}, vol. 261, p. 112147,
  2022.

\bibitem{c30}
P.~Sunehag, G.~Lever, A.~Gruslys, W.~M. Czarnecki, V.~Zambaldi, M.~Jaderberg,
  M.~Lanctot, N.~Sonnerat, J.~Z. Leibo, K.~Tuyls \emph{et~al.},
  ``Value-decomposition networks for cooperative multi-agent learning,''
  \emph{arXiv preprint arXiv:1706.05296}, 2017.

\bibitem{c31}
T.~Rashid, C.~De~Witt, G.~Farquhar, J.~Foerster, S.~Whiteson, and M.~Samvelyan,
  ``Qmix: Monotonic value function factorisation for deep multi-agent
  reinforcement learning,'' in \emph{35th International Conference on Machine
  Learning, ICML 2018}, pp. 6846--6859, 2018.

\bibitem{c32}
K.~Son, D.~Kim, W.~J. Kang, D.~E. Hostallero, and Y.~Yi, ``Qtran: Learning to
  factorize with transformation for cooperative multi-agent reinforcement
  learning,'' in \emph{International conference on machine learning}, pp.
  5887--5896.\hskip 1em plus 0.5em minus 0.4em\relax PMLR, 2019.

\bibitem{c33}
A.~Mahajan, T.~Rashid, M.~Samvelyan, and S.~Whiteson, ``Maven: Multi-agent
  variational exploration,'' \emph{Advances in Neural Information Processing
  Systems}, vol.~32, 2019.

\bibitem{c34}
Y.~Yang, J.~Hao, B.~Liao, K.~Shao, G.~Chen, W.~Liu, and H.~Tang, ``Qatten: A
  general framework for cooperative multiagent reinforcement learning,''
  \emph{arXiv preprint arXiv:2002.03939}, 2020.

\bibitem{c26}
P.~Peng, Y.~Wen, Y.~Yang, Q.~Yuan, Z.~Tang, H.~Long, and J.~Wang, ``Multiagent
  bidirectionally-coordinated nets: Emergence of human-level coordination in
  learning to play starcraft combat games,'' \emph{arXiv preprint
  arXiv:1703.10069}, 2017.

\bibitem{c27}
J.~Jiang and Z.~Lu, ``Learning attentional communication for multi-agent
  cooperation,'' \emph{Advances in neural information processing systems},
  vol.~31, 2018.

\bibitem{c28}
A.~Das, T.~Gervet, J.~Romoff, D.~Batra, D.~Parikh, M.~Rabbat, and J.~Pineau,
  ``Tarmac: Targeted multi-agent communication,'' in \emph{International
  Conference on Machine Learning}, pp. 1538--1546.\hskip 1em plus 0.5em minus
  0.4em\relax PMLR, 2019.

\bibitem{c29}
M.~Rangwala and R.~Williams, ``Learning multi-agent communication through
  structured attentive reasoning,'' \emph{Advances in Neural Information
  Processing Systems}, vol.~33, pp. 10\,088--10\,098, 2020.

\bibitem{c18}
R.~Lowe, Y.~I. Wu, A.~Tamar, J.~Harb, O.~Pieter~Abbeel, and I.~Mordatch,
  ``Multi-agent actor-critic for mixed cooperative-competitive environments,''
  \emph{Advances in neural information processing systems}, vol.~30, 2017.

\bibitem{c19}
S.~Iqbal and F.~Sha, ``Actor-attention-critic for multi-agent reinforcement
  learning,'' in \emph{International conference on machine learning}, pp.
  2961--2970.\hskip 1em plus 0.5em minus 0.4em\relax PMLR, 2019.

\bibitem{c25}
J.~Foerster, G.~Farquhar, T.~Afouras, N.~Nardelli, and S.~Whiteson,
  ``Counterfactual multi-agent policy gradients,'' in \emph{Proceedings of the
  AAAI conference on artificial intelligence}, vol.~32, no.~1, 2018.

\bibitem{c21}
C.~Yu, A.~Velu, E.~Vinitsky, J.~Gao, Y.~Wang, A.~Bayen, and Y.~Wu, ``The
  surprising effectiveness of ppo in cooperative multi-agent games,''
  \emph{Advances in Neural Information Processing Systems}, vol.~35, pp.
  24\,611--24\,624, 2022.

\bibitem{c12}
H.~Huang, W.~Zhang, J.~Ding, D.~M. Stipanovi{\'c}, and C.~J. Tomlin,
  ``Guaranteed decentralized pursuit-evasion in the plane with multiple
  pursuers,'' in \emph{2011 50th IEEE Conference on Decision and Control and
  European Control Conference}, pp. 4835--4840.\hskip 1em plus 0.5em minus
  0.4em\relax IEEE, 2011.

\bibitem{c1}
Z.~Zhou, W.~Zhang, J.~Ding, H.~Huang, D.~M. Stipanovi{\'c}, and C.~J. Tomlin,
  ``Cooperative pursuit with voronoi partitions,'' \emph{Automatica}, vol.~72,
  pp. 64--72, 2016.

\bibitem{c13}
A.~Pierson and D.~Rus, ``Distributed target tracking in cluttered environments
  with guaranteed collision avoidance,'' in \emph{2017 International Symposium
  on Multi-Robot and Multi-Agent Systems (MRS)}, pp. 83--89.\hskip 1em plus
  0.5em minus 0.4em\relax IEEE, 2017.

\bibitem{c11}
W.~L. Scott and N.~E. Leonard, ``Optimal evasive strategies for multiple
  interacting agents with motion constraints,'' \emph{Automatica}, vol.~94, pp.
  26--34, 2018.

\bibitem{c15}
B.~Tian, P.~Li, H.~Lu, Q.~Zong, and L.~He, ``Distributed pursuit of an evader
  with collision and obstacle avoidance,'' \emph{IEEE Transactions on
  Cybernetics}, vol.~52, no.~12, pp. 13\,512--13\,520, 2021.

\bibitem{c2}
J.~Chen, W.~Zha, Z.~Peng, and D.~Gu, ``Multi-player pursuit--evasion games with
  one superior evader,'' \emph{Automatica}, vol.~71, pp. 24--32, 2016.

\bibitem{c3}
L.~Angelani, ``Collective predation and escape strategies,'' \emph{Physical
  review letters}, vol. 109, no.~11, p. 118104, 2012.

\bibitem{c5}
T.~Vicsek, A.~Czir{\'o}k, E.~Ben-Jacob, I.~Cohen, and O.~Shochet, ``Novel type
  of phase transition in a system of self-driven particles,'' \emph{Physical
  review letters}, vol.~75, no.~6, p. 1226, 1995.

\bibitem{c4}
M.~Janosov, C.~Vir{\'a}gh, G.~V{\'a}s{\'a}rhelyi, and T.~Vicsek, ``Group
  chasing tactics: how to catch a faster prey,'' \emph{New Journal of Physics},
  vol.~19, no.~5, p. 053003, 2017.

\bibitem{c6}
M.~H{\"u}ttenrauch, S.~Adrian, G.~Neumann \emph{et~al.}, ``Deep reinforcement
  learning for swarm systems,'' \emph{Journal of Machine Learning Research},
  vol.~20, no.~54, pp. 1--31, 2019.

\bibitem{c7}
C.~De~Souza, R.~Newbury, A.~Cosgun, P.~Castillo, B.~Vidolov, and D.~Kuli{\'c},
  ``Decentralized multi-agent pursuit using deep reinforcement learning,''
  \emph{IEEE Robotics and Automation Letters}, vol.~6, no.~3, pp. 4552--4559,
  2021.

\bibitem{c8}
Z.~Zhang, X.~Wang, Q.~Zhang, and T.~Hu, ``Multi-robot cooperative pursuit via
  potential field-enhanced reinforcement learning,'' \emph{arXiv preprint
  arXiv:2203.04700}, 2022.

\bibitem{c9}
K.~Wan, D.~Wu, Y.~Zhai, B.~Li, X.~Gao, and Z.~Hu, ``An improved approach
  towards multi-agent pursuit--evasion game decision-making using deep
  reinforcement learning,'' \emph{Entropy}, vol.~23, no.~11, p. 1433, 2021.

\bibitem{c10}
R.~Zhang, Q.~Zong, X.~Zhang, L.~Dou, and B.~Tian, ``Game of drones: Multi-uav
  pursuit-evasion game with online motion planning by deep reinforcement
  learning,'' \emph{IEEE Transactions on Neural Networks and Learning Systems},
  2022.

\bibitem{c17}
J.~A. Sethian, ``Fast marching methods,'' \emph{SIAM review}, vol.~41, no.~2,
  pp. 199--235, 1999.

\bibitem{c16}
F.~Xia, A.~R. Zamir, Z.~He, A.~Sax, J.~Malik, and S.~Savarese, ``Gibson env:
  Real-world perception for embodied agents,'' in \emph{Proceedings of the IEEE
  conference on computer vision and pattern recognition}, pp. 9068--9079, 2018.

\bibitem{c20}
C.~S. de~Witt, T.~Gupta, D.~Makoviichuk, V.~Makoviychuk, P.~H. Torr, M.~Sun,
  and S.~Whiteson, ``Is independent learning all you need in the starcraft
  multi-agent challenge?'' \emph{arXiv preprint arXiv:2011.09533}, 2020.

\end{thebibliography}

\end{document}